\documentclass[9 pt,journal]{IEEEtran}

\usepackage{graphicx}
\usepackage{hyperref}
\usepackage[cmex10]{amsmath}
\usepackage{amssymb}
\usepackage{url}
\usepackage{titlesec}

\usepackage{subcaption}  
\usepackage{enumitem}  

\usepackage{array}
\usepackage{adjustbox}

\usepackage{url}
\usepackage[noadjust]{cite}
\usepackage{color}

\usepackage[noadjust]{cite}
\usepackage{color}

\usepackage{algorithm}
\usepackage{algpseudocode}

\usepackage{amssymb}
\usepackage{nomencl}
\makenomenclature

\usepackage{etoolbox}

\titlespacing*{\section}{0pt}{\baselineskip}{0pt}
\titlespacing*{\subsection}{-20pt}{\baselineskip}{-20pt}
\titlespacing*{\subsubsection}{0pt}{\baselineskip}{0pt}
\titlespacing*{\subsection}{-20pt}{-30pt}{-20pt}

\begin{document}

\title{Scalable Volt-VAR Optimization using RLlib-IMPALA Framework: A Reinforcement Learning Approach}

\author{{Alaa Selim, \emph{Student Member, IEEE}, Yanzhu Ye, \emph{Member, IEEE}, Junbo Zhao, \emph{Senior Member, IEEE}, Bo Yang, \emph{Senior Member, IEEE} 

\thanks{
A. Selim and J. Zhao  are with the Department of Electrical and Computer Engineering, University of Connecticut (e-mail: alaa.selim@uconn.edu, junbo@uconn.edu).
Y. Ye and B.Yang are with the energy solution lab, Hitachi America Ltd R\&D (e-mail: Yanzhu.Ye@hal.hitachi.com , Bo.Yang@hal.hitachi.com).
}

}}

\markboth{}%
{Shell \MakeLowercase{\textit{et al.}}: Bare Demo of IEEEtran.cls for Journals}

\maketitle


\begin{abstract}

In the rapidly evolving domain of electrical power systems, the Volt-VAR optimization (VVO) is increasingly critical, especially with the burgeoning integration of renewable energy sources. Traditional approaches to learning-based VVO in expansive and dynamically changing power systems are often hindered by computational complexities. To address this challenge, our research presents a novel framework that harnesses the potential of Deep Reinforcement Learning (DRL), specifically utilizing the Importance Weighted Actor-Learner Architecture (IMPALA) algorithm, executed on the RAY platform. This framework, built upon RLlib—an industry-standard in Reinforcement Learning—ingeniously capitalizes on the distributed computing capabilities and advanced hyperparameter tuning offered by RAY. This design significantly expedites the exploration and exploitation phases in the VVO solution space. Our empirical results demonstrate that our approach not only surpasses existing DRL methods in achieving superior reward outcomes but also manifests a remarkable tenfold reduction in computational requirements. The integration of our DRL agent with the RAY platform facilitates the creation of RLlib-IMPALA, a novel framework that efficiently uses RAY's resources to improve system adaptability and control. RLlib-IMPALA leverages RAY's toolkit to enhance analytical capabilities and significantly speeds up training to become more than 10 times faster than other state-of-the-art DRL methods. 
\end{abstract}
\vspace{-0.1cm}

\vspace{-0.3cm}
\section{Introduction}
\IEEEPARstart{L}{earning}-based Volt-VAR control faces a significant challenge in training time, especially for high-dimensional problems. The complexity and high-dimensionality of these control problems necessitate the use of sophisticated machine learning and optimization techniques for efficient solutions. The duration and effectiveness of the training phase are critical, as they directly impact the performance and reliability of Volt-VAR control algorithms in managing voltage profiles and reactive power in power distribution systems. 

As we navigate through the corpus of existing literature, it becomes evident that Deep Reinforcement Learning (DRL) has been instrumental in addressing the intricate high-dimensionality challenges presented by Volt-VAR optimization (VVO) \cite{cao2020reinforcement}. Our review focuses on illuminating a selection of these studies, particularly highlighting the duration of DRL training required to surmount these complexities. Traditional model-based and heuristic algorithms are becoming increasingly incompetent due to slow online calculation speed \cite{ma2023simplified}. In contrast, DRL has been recognized as an effective alternative, although it requires training which could be time-consuming given the complexity of the problem \cite{fan2022powergym}. In \cite{ma2023simplified}, the paper introduces methods to optimize DRL for VVO probelm, including simplifying critic network training by directly setting the reward function as the action value, deferring actor network training until post-critic network maturation, and implementing side-tuning transfer learning for quick adaptation to new power system topologies. These strategies collectively enhance the DRL process, significantly reducing training complexity and time, potentially to a range of approximately 3 to 40 minutes. In \cite{liu2022bi}, while 100 and 500 episodes are deemed sufficient for training DRL on 33-bus and 123-bus cases respectively, the absence of wall-clock training times leaves the scalability and practicality of these DRL methods for high-dimensional problems unquantified, a critical oversight when evaluating real-world applicability. Another research in \cite{gupta2023deep}, the authors address long training durations and convergence challenges in DRL models, applying Nesterov’s accelerated methods to simplify RNN architectures. Despite extended training periods, these methods reduce costs by over 50\%, improving voltage profiles. This highlights a strategic compromise between computational effort and performance, suggesting advanced optimization as a potential solution to these computational hurdles. Literature insufficiently tackles the challenge of lengthy training times in DRL models, underscoring a critical area for enhanced understanding and improvement.

The burden of training time in DRL for high-dimensional action spaces is a well-documented issue across various domains. This burden arises from the complexity and the vastness of the action spaces that need to be explored during the training process. In a study related to developing high-dimensional action spaces, the paper \cite{manna2022learning} introduces a novel Reinforcement Learning (RL) strategy that combines decision trees with modified rewards, efficient sampling, and a window scaling scheme. This approach is designed to make the exploration and exploitation of continuous action spaces more efficient and scalable, significantly reducing the time required for such tasks in materials science. 

Another paper \cite{ming2023cooperative} discussed cooperative modular reinforcement learning for large discrete action spaces mentions that DRL has achieved remarkable results on high-dimension state tasks. However, it suffers from hard convergence and low sample efficiency when solving large discrete action space problems, indicating the training time burden and the need for more efficient training algorithms. In the domain of trajectory planning, RL has demonstrated potential in solving complex control, guidance, and coordination problems but can suffer from long training times and poor performance. The paper \cite{williams2022trajectory} suggests that using high-level actions can reduce the required number of training steps, thereby potentially reducing training time and enhancing the performance of an RL-trained guidance policy.

Another research \cite{kanervisto2020action} on action space shaping in DRL highlights the practice of modifying and shrinking the action space from its original size to avoid "pointless" actions and ease implementation. This action space manipulation is mostly done based on intuition, with little systematic research supporting the design, indicating a potential area for reducing training time through more systematic action space shaping. These papers elucidate the challenges and potential strategies to address the training time burden in RL for high-dimensional action spaces, demonstrating the multifaceted nature of this issue across different domains.

In this study, we pioneer an innovative framework, employing DRL strategies via the RAY platform \cite{raylibraries}, that facilitates accelerated convergence and training speedup, essential for handling large-scale computations in VVC problems. 
RAY's distributed framework support highly distributed RL workloads, abstracting away the complexities of distributed computing, which is pivotal for developing and deploying large-scale AI models efficiently \cite{anyscale}. The major contributions are highlighted as follows:
\begin{enumerate}[topsep=-10pt]  
    \item \textbf{Optimal DERs Placement :}
    Introducing a novel algorithm for the strategic placement of Distributed Energy Resources (DERs), specifically solar photovoltaic (PV) systems and batteries, this study represents an initial step in the intricate solution of VVO. The approach is geared towards pinpointing the optimal location of solar PV installations to ensure minimal voltage violations, setting the stage for a subsequent DRL control solution.
    \item \textbf{Enhanced Centralized Control:}
    Utilizing the RAY platform for distributed computing, we introduce a robust centralized control agent built on the Importance Weighted Actor-Learner Architecture (IMPALA) \cite{espeholt2018impala}. This approach ensures scalability and efficient management of a high-dimensional action space. As a result, the agent adeptly mitigates voltage violations, especially in high DER penetration scenarios, while seamlessly handling both smart inverters and utility devices with continuous and discrete controls. 
    \item \textbf{Comparative Performance Analysis:}
   Through a comparative analysis with state-of-art methods like Soft Actor-Critic (SAC) \cite{haarnoja2018soft} and Proximal Policy Optimization (PPO) \cite{schulman2017proximal}, our centralized control approach, bolstered by distributed computing, emerges with superior performance. This combination ensures enhanced efficacy in assuring voltage regulation.
\end{enumerate}
\vspace{10pt}
The remainder of this paper is structured as follows: Section II expounds on the problem formulation and presents a proposed model for optimal planning of DERs concerning this problem. Section III unveils the proposed approach towards resolving the Volt-VAR problem, emphasizing the reduction in computation time. Subsequently, Section IV divulges the numerical results, underscoring the solid framework of the proposed approach. Lastly, Section V encapsulates the conclusions drawn from the study and delineates the avenues for future work in this domain.
\section{Problem Formualtion}
\begin{figure}[h]
  \centering
 \includegraphics[width=0.9\columnwidth]{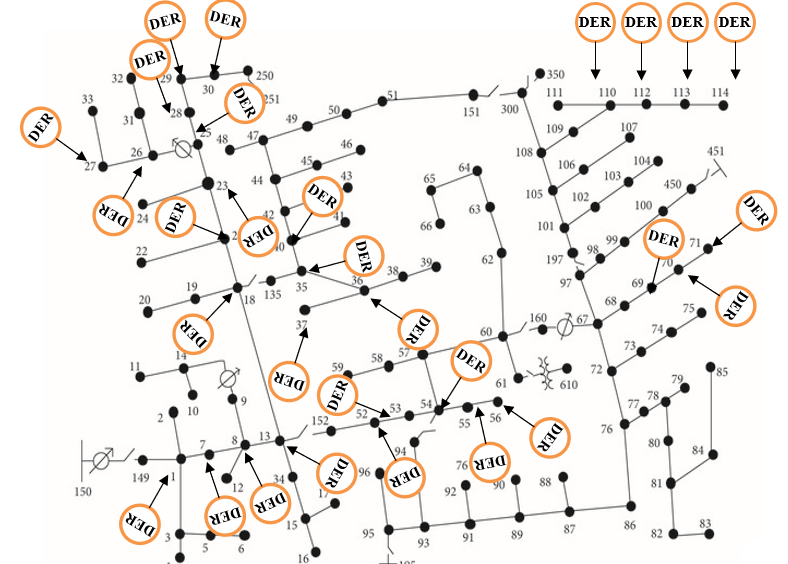 }
  \centering
  \caption{Optimal planning of DERs within the IEEE 123 testing feeder}
  \centering
  \label{Fig_Optimal planning of DERs within the distribution network}
\end{figure}

The objective is to minimize the total number of control steps and voltage violations over the time horizon \( T \) for the VVO problem as shown below: 
{\setlength\abovedisplayskip{0pt}
\setlength\belowdisplayskip{0pt}
\begin{equation}
\begin{aligned}
& \underset{Q_{\text{PV}}, P_{\text{PV}}, P_{\text{Batt}}, S_{\text{Cap}}, T_{\text{Tap}}}{\text{minimize}}
& & J = \alpha \sum_{t=1}^{T} N_{\text{con}}(t) + \beta \sum_{t=1}^{T} V_{\text{vio}}(t) \\
\end{aligned}
\end{equation}
}
Subject to:
\begin{align}
& Q_{\text{PV,min}} \leq Q_{\text{PV}}(t, n) \leq Q_{\text{PV,max}}, \quad \forall t \in T, \, \forall n \in N_{\text{PV}}, \\
& P_{\text{PV,min}} \leq P_{\text{PV}}(t, n) \leq P_{\text{PV,max}}, \quad \forall t \in T, \, \forall n \in N_{\text{PV}}, \\
& P_{\text{Batt,min}} \leq P_{\text{Batt}}(t, n) \leq P_{\text{Batt,max}}, \quad \forall t \in T, \, \forall n \in N_{\text{Batt}}, \\
& S_{\text{Cap}}(t, n) \in \{0, 1\}, \quad \forall t \in T, \, \forall n \in N_{\text{Cap}}, \\
& T_{\text{Tap,min}} \leq T_{\text{Tap}}(t,n) \leq T_{\text{Tap,max}}, \quad \forall t \in T, \forall n \in N_{\text{Trans}}, \\
& P(i,t) - \sum_{j \in \text{Buses}} G_{ij} V(i,t) V(j,t) \\
& \quad \times \cos(\theta(i,t) - \theta(j,t)) = 0, \quad \forall i \in \text{Buses}, \, \forall t \in T, \\
& Q(i,t) - \sum_{j \in \text{Buses}} G_{ij} V(i,t) V(j,t) \\
& \quad \times \sin(\theta(i,t) - \theta(j,t)) = 0, \quad \forall i \in \text{Buses}, \, \forall t \in T, \\
& V(i,t) V(j,t) \cos(\theta(i,t) - \theta(j,t)) \\
& \quad - P(i,j,t) = 0, \quad \forall i,j \in \text{Buses}, \, \forall t \in T, \\
& V(i,t) V(j,t) \sin(\theta(i,t) - \theta(j,t)) \\
& \quad - Q(i,j,t) = 0, \quad \forall i,j \in \text{Buses}, \, \forall t \in T.
\end{align}
Here, \( \alpha \) and \( \beta \) are the weights assigned to \( N_{\text{con}}(t) \), the number of control steps of DERs and utility devices  and the number of voltage violations \( V_{\text{vio}}(t) \), respectively, at each time step \( t \).
The problem is bounded by several operational constraints to ensure system stability and adherence to technical specifications of Volt-VAR problem. The active and reactive power output of PV systems and battery storage are bounded within \( P_{\text{PV,min}} \), \( P_{\text{PV,max}} \), \( Q_{\text{PV,min}} \), \( Q_{\text{PV,max}} \), \( P_{\text{Batt,min}} \), and \( P_{\text{Batt,max}} \) respectively. The shunt capacitor status is represented as a binary variable \( S_{\text{Cap}}(t) \), and the transformer tap settings are bounded within \( T_{\text{Tap,min}} \) and \( T_{\text{Tap,max}} \). The power flow equations ensure the conservation of active and reactive power in the system. 
 Algorithm 1 strategically identifies optimal bus locations for the placement of PV systems and batteries within a power distribution network. It operates by assessing each bus's suitability based on a year-long hourly simulation that accounts for varying irradiance levels. The algorithm quantifies the fitness of each bus based on the cumulative annual voltage violations and power losses, essential criteria for stable network operation and ranks the buses for the least cost function. As shown in Fig. \ref{Fig_Optimal planning of DERs within the distribution network}, 30 pairs of PV systems and batteries are added onto the IEEE 123 test feeder based on the ranking obtained from the fitness function from the algorithm,  to use their current locations for solving the VVO problem.
\begin{algorithm}
\caption{Optimal Bus Placement for PV and Battery in a Power Network}
\begin{algorithmic}[1]
\State \textbf{Input:} Set of buses \( B \), Irradiance profile \( I \)
\State \textbf{Output:} Ranked bus pairs for colocated PV and battery placement

\Procedure{EvaluateFitness}{$pv\_bus, battery\_bus$}
    \State \( V_{total} \leftarrow 0 \) 
    \State \( L_{total} \leftarrow 0 \) 
    \For{\( hour \) in \( 1 \) to \( 8760 \)}
        \State \( P_{active} \leftarrow I_{hour} \times 100 \)
        \State Set PV \( Pmpp \) to \( P_{active} \)
        \State Solve power flow for \( hour \)
        \State \( V_{violations} \leftarrow \) count of voltages outside \( [0.95, 1.05] \)
        \State \( V_{total} \leftarrow V_{total} + V_{violations} \)
        \State \( L_{total} \leftarrow L_{total} + \) power loss for \( hour \)
    \EndFor
    \State \textbf{return} \( V_{total} + L_{total} \)
\EndProcedure

\State Load \( I \) from file
\State Initialize power network
\State Define PV and Storage
\State \( R \leftarrow \) empty list
\For{bus in \( B \)}
    \State \( fitness \leftarrow \) \Call{EvaluateFitness}{bus, bus}
    \State Append \( (bus, bus, fitness) \) to \( R \)
\EndFor
\State Sort \( R \) by \( fitness \)
\State \textbf{print} ranked bus pairs from \( R \)
\label{algorithm1}
\end{algorithmic}
\end{algorithm}
\section{Proposed IMPALA-DRL Framework}
\begin{figure*}
\centering
\includegraphics[width=0.8\linewidth]{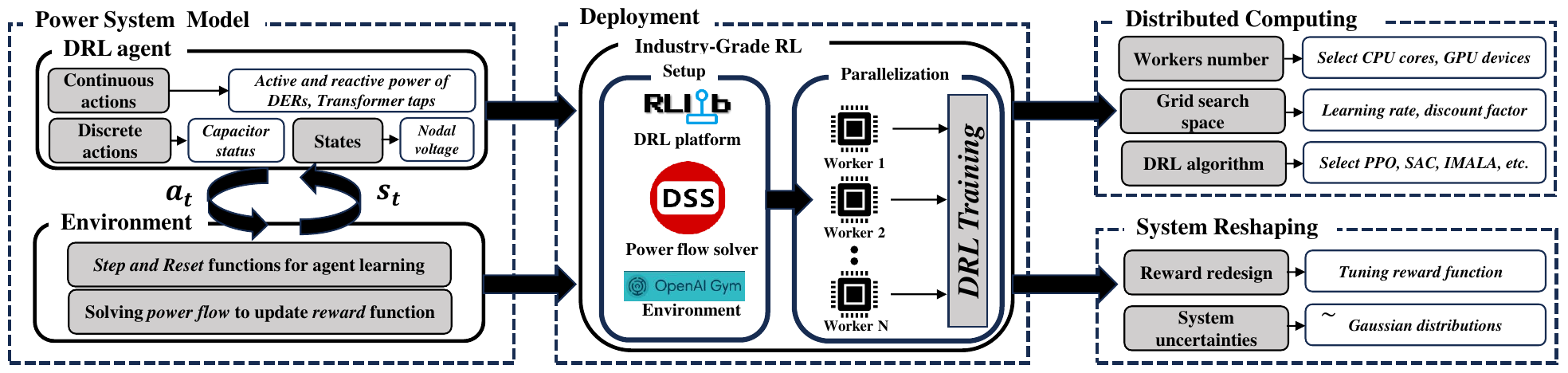}
\caption{Proposed framework for  RLlib-IMPALA}
 \label{Proposed framework}
\end{figure*}

The Importance Weighted Actor-Learner Architecture (\textbf{IMPALA}) algorithm \cite{espeholt2018impala} revolutionizes DRL by introducing a scalable and efficient framework specifically designed for parallel environments. This innovative method distinctively separates the roles of actors and learners during the learning process, which significantly enhances both stability and sample efficiency. In the IMPALA framework, actors are responsible for interacting with their individual environments and generating trajectories, which are sequences of states, actions, and rewards, denoted as $\{(s_t, a_t, r_t)\}_{t=1}^n$. These trajectories are then sent to the central learner for processing. One of the pivotal aspects of IMPALA is its utilization of \textit{off-policy} learning combined with \textit{importance sampling}. This approach effectively addresses the discrepancies between the actors' policies, denoted as $\mu$, and the central policy $\pi$ maintained by the learner. In the context of VVO, IMPALA's capability to manage high-dimensional control spaces, both continuous and discrete, becomes crucial. It allows for fine-tuned control of various DERs, optimizing their operation for voltage stability. Furthermore, IMPALA's distributed nature speeds up the training process, essential for handling the complexities of the distribution network. The algorithm operates iteratively and involves several key computational steps, which are enumerated below:

\begin{enumerate}
    \item The actors generate trajectories based on their local policies and forward them to the learner. This process is represented as:
    \begin{equation}
    \{(s_t, a_t, r_t)\}_{t=1}^n \sim \mu(.|s_t).
    \end{equation}
    Here, $\mu(.|s_t)$ represents the policy used by the actor, which is a probability distribution over the actions given the current state. Specifically for VVO applications, IMPALA's capacity to explore a wide range of actions through its actors, rather than restricting to a single trajectory, is particularly beneficial. This expansive exploration strategy enables faster learning and a more comprehensive identification of optimal control strategies for high-dimensional DERs, a critical feature for effective VVO.

    \item The importance sampling ratio, crucial for adjusting the weight of each sample, is computed as follows:
    \begin{equation}
    \rho_t = \min\left(\bar{\rho}, \frac{\pi(a_t|s_t)}{\mu(a_t|s_t)}\right),
    \end{equation}
    where $\bar{\rho}$ is a predefined clipping threshold that prevents the ratio from causing excessive variance in the updates. The term $\pi(a_t|s_t)$ denotes the probability of taking action $a_t$ under the central policy $\pi$ given the state $s_t$, and $\mu(a_t|s_t)$ is the corresponding probability under the actor's policy. This sampling ratio in IMPALA plays a key role in preventing biases or variances towards aggressive VVO actions and getting stuck in sub-optimal values.

 \item The learner calculates the value estimates by applying the Bellman equation with an importance sampling correction. The corrected value estimate is given by:
\begin{equation}
v_s = V(s_m) + \sum_{t=m}^{m+n-1} \gamma^{t-m} \left( \prod_{i=m}^{t-1} c_i \right) \delta_t^V,
\end{equation}
where $\delta_t^V = \rho_t \left( r_t + \gamma V(s_{t+1}) - V(s_t) \right)$, and $c_i = \min\left(\bar{c}, \frac{\pi(a_i|s_i)}{\mu(a_i|s_i)}\right)$. Here, $\gamma$ is the discount factor, and $\delta_t^V$ represents the temporal difference error adjusted by the importance sampling ratio. The equation crucially enables the IMPALA algorithm to effectively evaluate and adapt VVC strategies, considering both immediate rewards and long-term impacts. By incorporating importance sampling, it ensures stable learning across varied control scenarios in power systems. This approach enhances real-time response and optimizes long-term system stability and efficiency in VVC applications.
\item The policy gradient and the value function updates are performed using the following expressions. The policy is updated by maximizing the expected return, adjusted by the importance sampling ratio:
\begin{align}
    &\rho_s \nabla \log \pi(a_m | s_m) \left( r_m + \gamma v_{m+1} - V(s_m) \right) \nonumber \\
    &\quad - \nabla \sum_a \pi(a | s_m) \log \pi(a | s_m),
\end{align}
and the value function is updated to minimize the prediction error:
\begin{equation}
\nabla (v_m - V(s_m))^2.
\end{equation}
 For VVC, these updates are essential in refining the control policies for DERs, ensuring that the resulting actions effectively contribute to maintaining optimal voltage levels and reactive power balance within the power distribution network.
\end{enumerate}
To apply IMPALA as shown in Fig. \ref{Proposed framework}, the state \(s_t\), actions \(a_t\), and rewards are defined to capture the power system's dynamics at time step \(t\). The state \(s_t = [V_1, \ldots, V_N, D_1, \ldots, D_M]^{\top}\) includes per-unit voltage magnitudes \(V_i\) at each bus \(i\) and load demands \(D_j\). The action \(a_t\) involves controls for PV systems, batteries, transformer taps, and capacitors, represented as \(a_t = [Q_{\text{PV},1}, \ldots, P_{\text{PV},N}, \ldots, P_{\text{Batt},N}, \ldots, S_{cap},\ldots T_{Tap}]^{\top}\). The reward function \( r(s_t, a_t) \) aims to minimize voltage violations, formulated as \( r(s_t, a_t) = -  V_{\text{vio}} \). This reward function is defined based on (1), but we do not not include control steps in reward to limit each experiment to have one control step, which is the best global solution for \( N_{\text{con}}(t) \) to achieve.




\section{Numerical results}
In the numerical analysis section, we assess the impact of high solar PV penetration on the IEEE 123-bus system, using solar irradiance based on conditions in Santa Clara, California, and load profiles with nodes' interactions modeled via a random Gaussian distribution. Computations are performed on a PC system with an Intel(R) Core(TM) i9-12900KF 3.19GHz CPU, 128GB RAM, NVIDIA GeForce RTX 3090 24GB GPU, and 24 CPU cores, significantly enhancing simulation efficiency. In this setup, we control 30 utility-owned solar PVs with a total of 60 setpoints for active and reactive power, alongside the active power of 30 utility-owned batteries. These PVs and batteries are optimally placed following the Algorithm 1, in addition to managing 4 capacitor switching operations and the main transformer taps. Utilizing our proposed framework of \href{https://github.com/AlaaSelim-sudo/-RLlib-IMPALA}{RLlib-IMPALA (available on GitHub)} shown in Fig. \ref{Proposed framework}, the default FIFO (First In, First Out) scheduling algorithm is employed, promoting efficient resource allocation and execution transparency through the OpenDSS \cite{montenegro2012real} software (the power flow solver). The configurations for the IMPALA algorithm are elaborated in Table I. To ensure a fair comparison between IMPALA, PPO, and SAC algorithms, a methodical tuning process is undertaken. For PPO and IMPALA, common hyperparameters are set through \texttt{tune.grid\_search} with \texttt{lr=0.0005}, \texttt{train\_batch\_size=2500}, and \texttt{minibatch\_buffer\_size=128}. On the other hand, due to SAC's slower convergence, a learning rate of 0.003 and a training batch size of 256 are employed to expedite its convergence, ensuring a balanced assessment of each algorithm's performance in addressing VVO convergence challenges.
\begin{table}[H]
\caption{IMPALA Configuration Settings}
\centering
\begin{tabular}{lc}
\hline
\textbf{Parameter} & \textbf{Value} \\
\hline
V-trace & Enabled \\
Num SGD Iterations & 1 \\
Learner Queue Size & 16 \\
Grad Clip & 40.0 \\
Optimizer Type & Adam \\
Learning Rate & 0.0005 \\
Entropy Coefficient & 0.01 \\
Train Batch Size & 2500 \\
Num Rollout Workers & varies \\
Rollout Fragment Length & 50 \\
Exploration Config & Stochastic Sampling \\
\hline
\end{tabular}
\end{table}

\begin{figure}[H]
\centering
\includegraphics[width=0.7\linewidth]{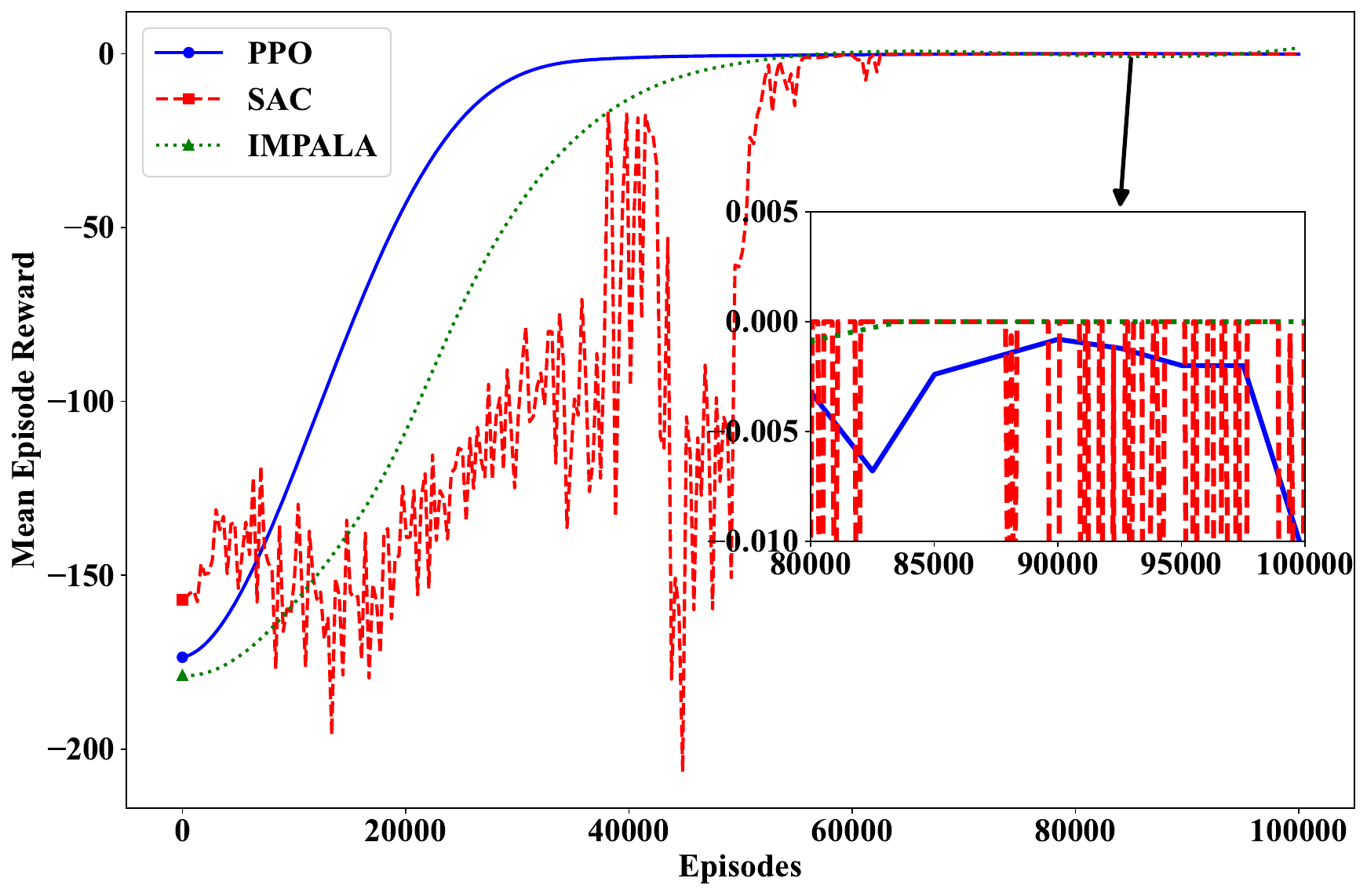}
\caption{Evaluation of the the learning curves of DRL agents on RAY }
\label{Evaluation of the the learning curves }
\end{figure}

\begin{figure*}[!htbp]  
    \centering
    \begin{subfigure}[b]{0.34\textwidth}
        \centering
        \includegraphics[width=\textwidth]{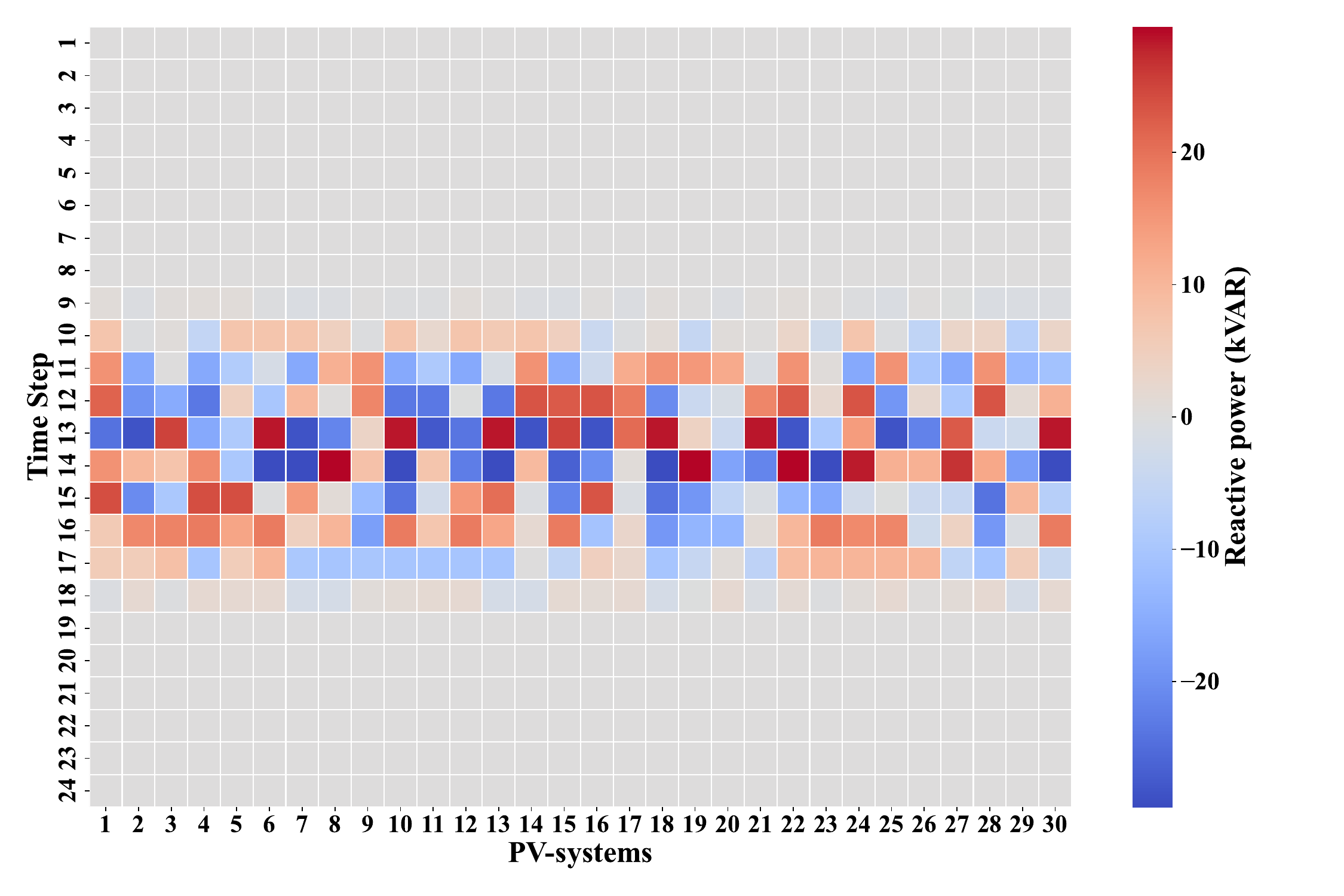}
        \caption{Evaluation of PV-kVAR setpoints}
        \label{fig:pv_kvar}
    \end{subfigure}%
    \begin{subfigure}[b]{0.34\textwidth}
        \centering
        \includegraphics[width=\textwidth]{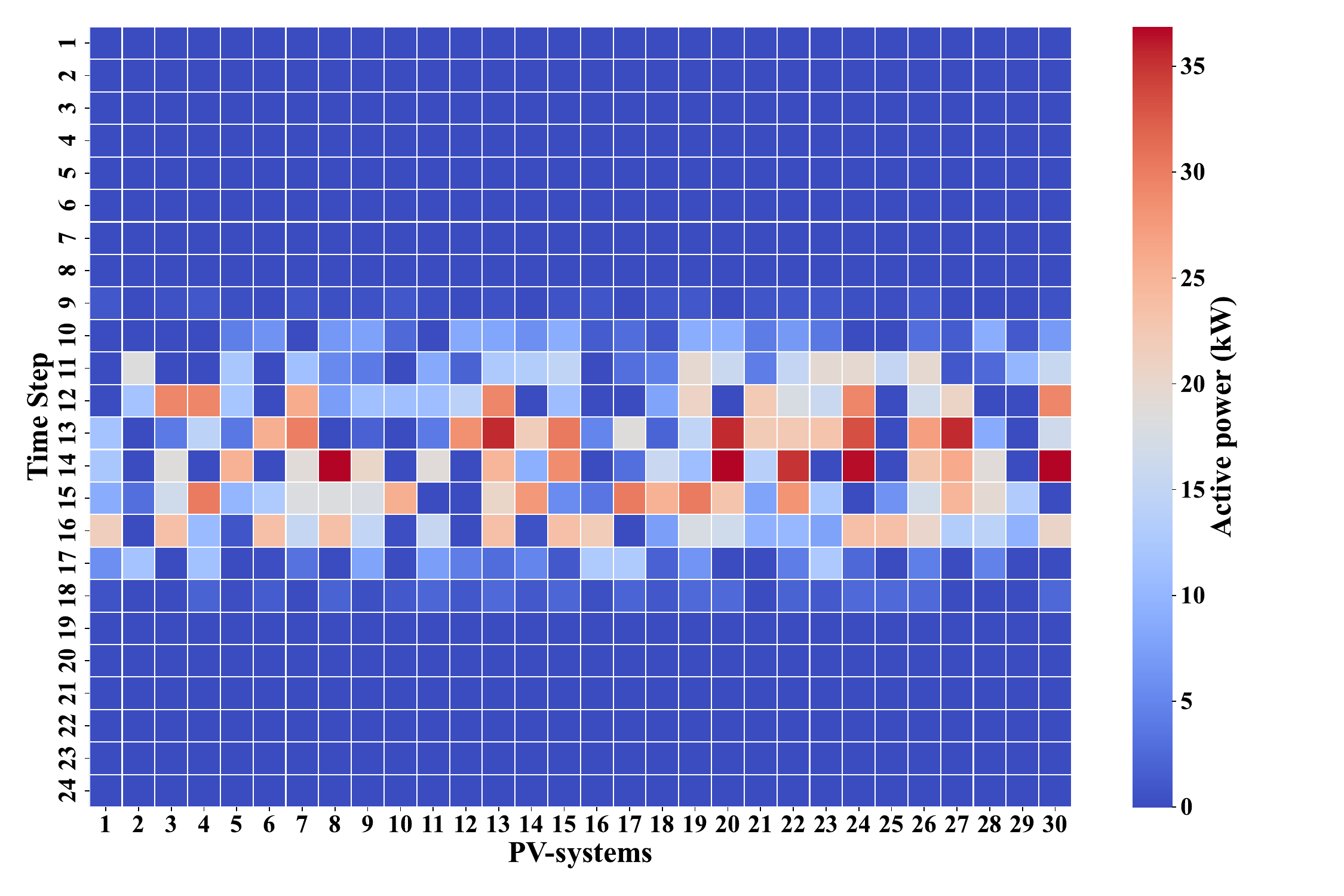}
        \caption{Evaluation of PV-kW setpoints}
        \label{fig:pv_kw}
    \end{subfigure}%
    \begin{subfigure}[b]{0.34\textwidth}
        \centering
        \includegraphics[width=\textwidth]{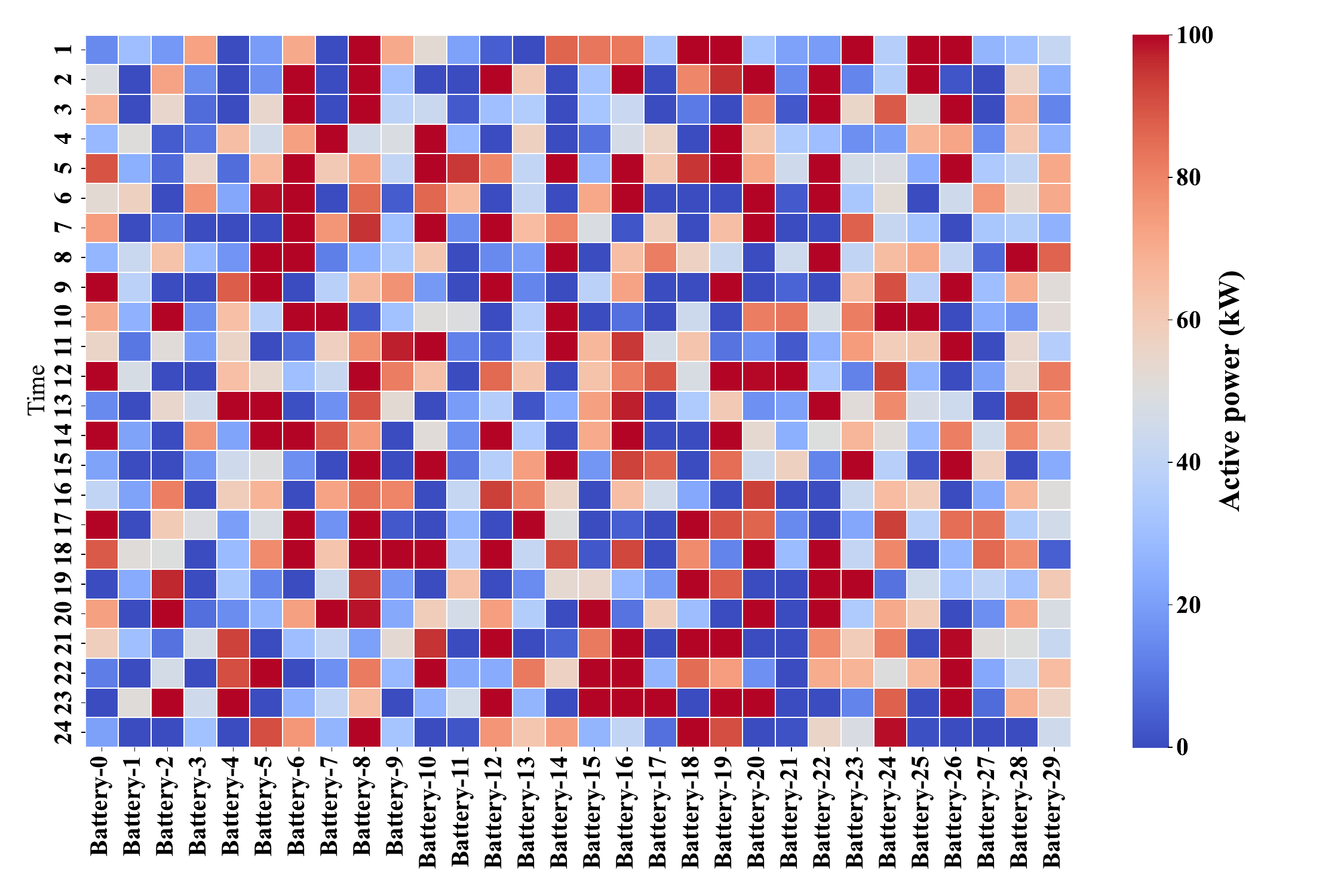}
        \caption{Evaluation of Batteries' setpoints}
        \label{fig:battery_kw}
    \end{subfigure}
    \caption{Evaluations of the RLlib-IMPALA on controlling DER setpoints}
    \label{fig:set_points_eval}
\end{figure*}

\begin{figure*}[!htbp]
    \centering
    \begin{subfigure}[b]{0.32\textwidth}
        \centering
        \includegraphics[width=\textwidth]{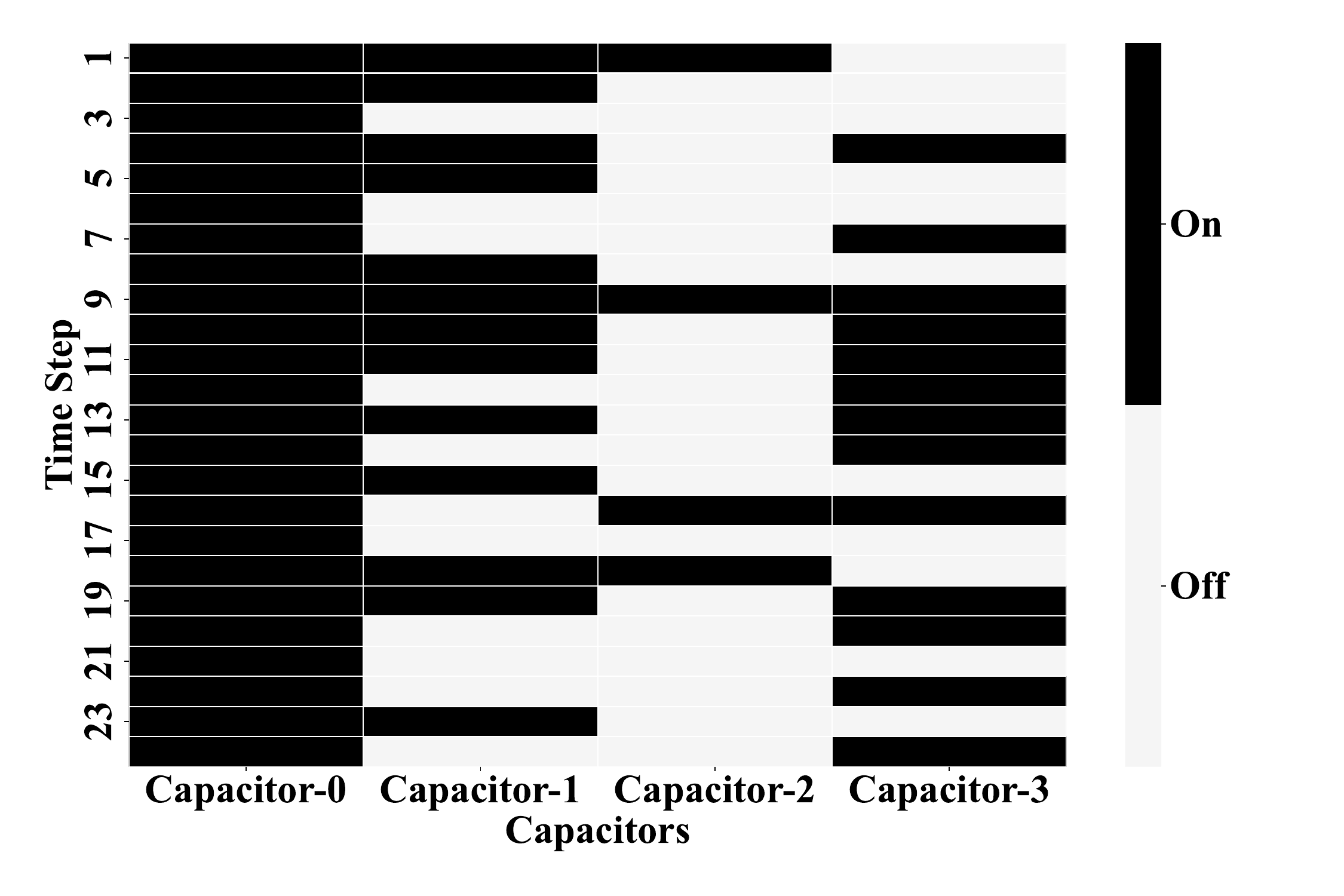}
        \caption{Capacitors' status}
        \label{fig:capacitor_status}
    \end{subfigure}%
    \begin{subfigure}[b]{0.33\textwidth}
        \centering
        \includegraphics[width=\textwidth]{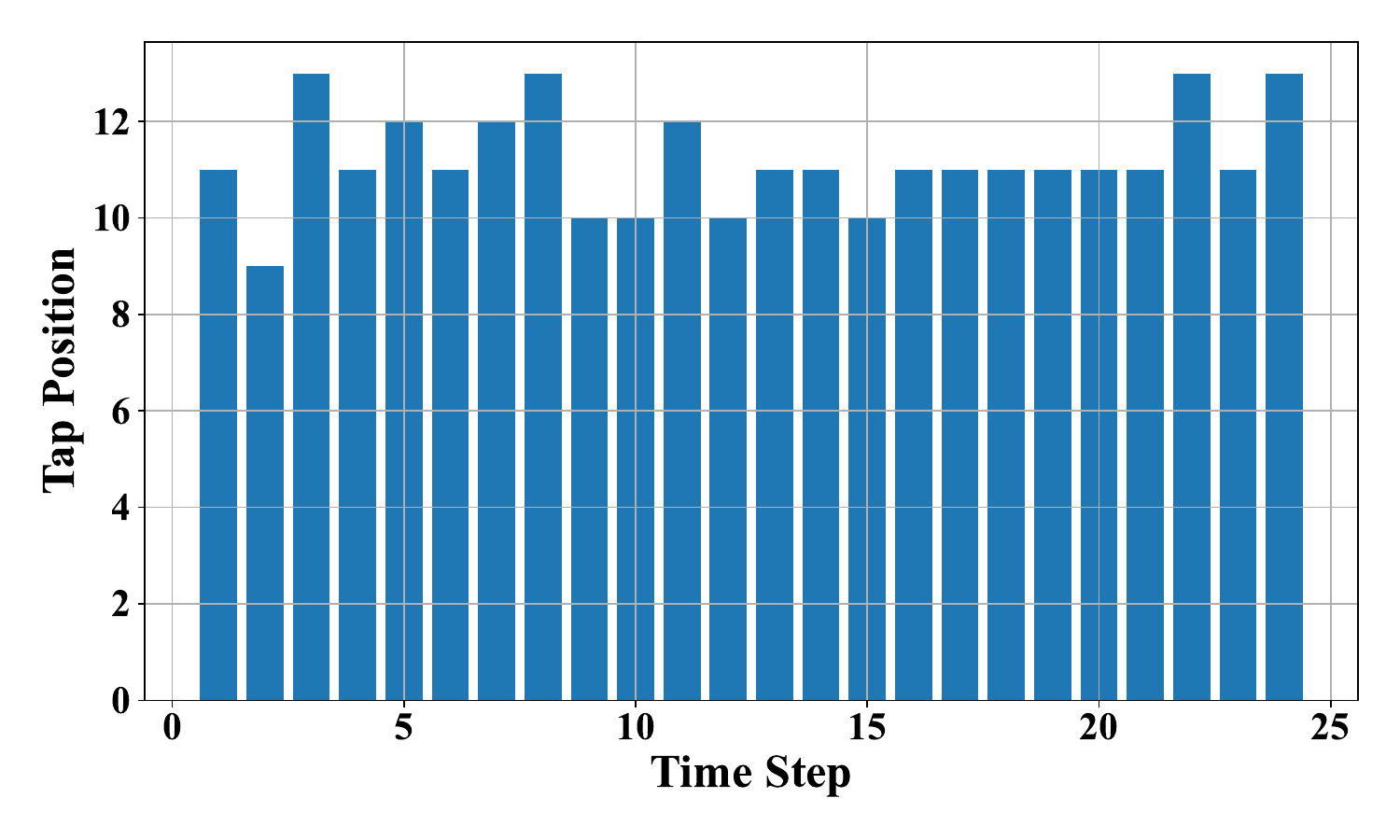}
        \caption{Transformer taps}
        \label{fig:transformer_status}
    \end{subfigure}
    \begin{subfigure}[b]{0.34\textwidth}
        \centering
        \includegraphics[width=\textwidth]{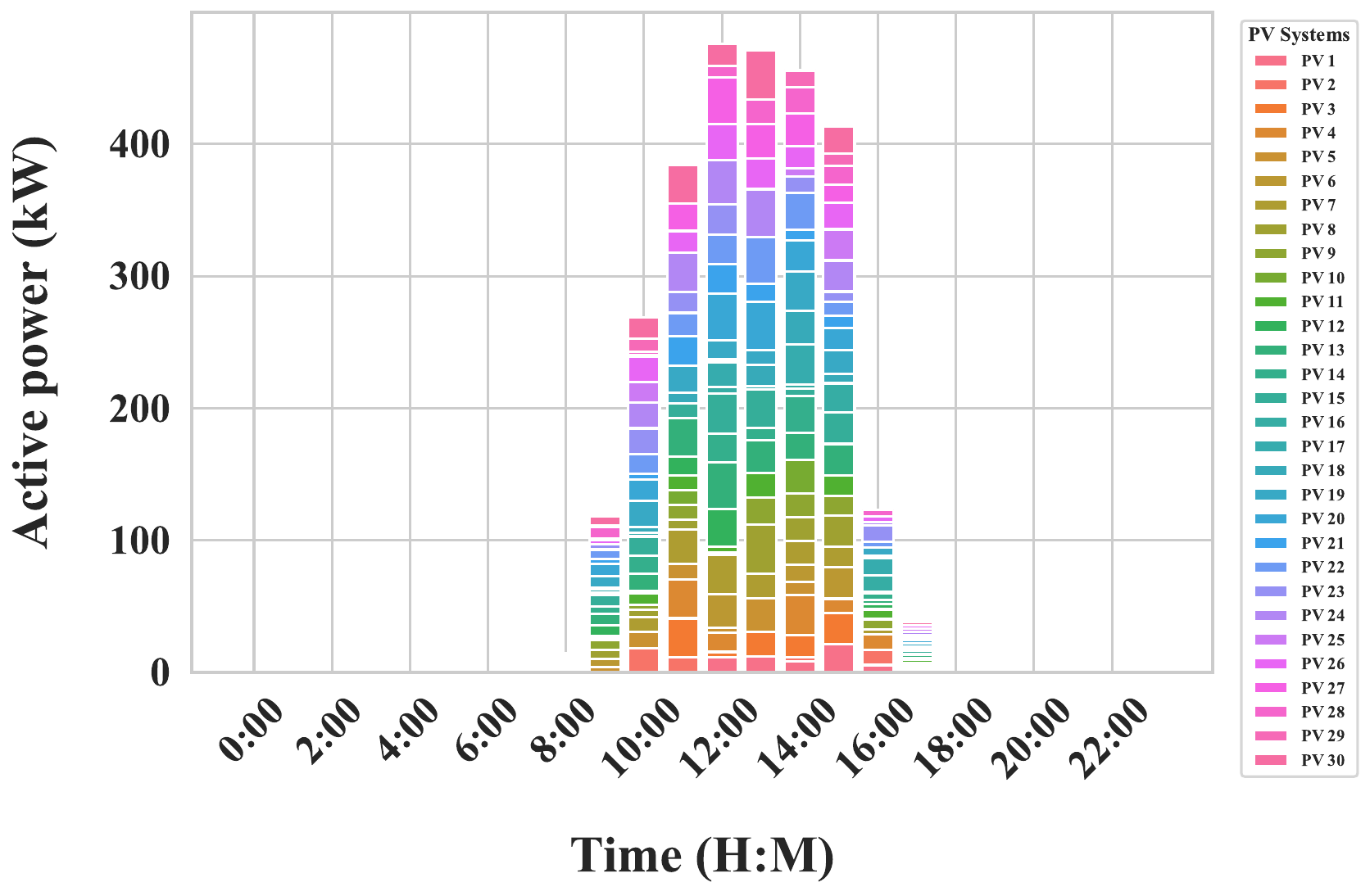}
        \caption{Evaluation of the active power injection }
        \label{fig:pv_power_injection}
    \end{subfigure}
    \caption{Evaluations of the RLlib-IMPALA on controlling capacitors, transformers and active power injection}
    \label{fig:system_components_eval1}
\end{figure*}

PPO exhibits a faster rate of convergence among the algorithms; however, IMPALA not only converges swiftly but also attains better stability in the reward value at the end of training as shown in Fig. \ref{Evaluation of the the learning curves }. Unlike SAC and PPO, which do not stabilize at a zero reward value equivalent to zero voltage violations---our desired target---IMPALA achieves this with the fastest training time among the other agents as shown in Table II. IMPALA's V-trace off-policy correction technique enhances learning stability and effectiveness from older data, whereas SAC necessitates more complex sampling techniques and softer value function update strategy, leading to slower convergence and demanding greater computational resources and training time.
Following the evaluation of the training time and computational capability of the proposed approach utilizing IMPALA, the voltage violations across a year-long profile for the Santa Clara location are assessed by testing random time steps for any hour of the year. 
Additionally, an evaluation of all setpoints over a one-day analysis is conducted, the results of which are illustrated in Figs. \ref{fig:set_points_eval} and \ref{fig:system_components_eval1}. These figures elucidate the real setpoints required by DERs, utility transformer and capacitors to achieve this high performance of zero voltage violations. Fig. \ref{fig:set_points_eval} illustrates the execution of setpoints for each PV system and batteries over a 24-hour horizon using IMAPLA. This figure illustrates the PV-kW and PV-kVAR setpoints, after considering the impact of irradiance, thereby emulating a realistic PV production curve. Additionally, the figure includes plot of dispensing active power from battery storage based on the operational constraint of the distribution system to have zero voltage violations. Fig. \ref{fig:system_components_eval1}.a illustrates the switching operations of four capacitors throughout the day. It clearly shows that Capacitor-0 remains "On" at all times, while the others are dynamically switched according to the system's requirements. Fig. \ref{fig:system_components_eval1}.b illustrates the discrete transformer tap positions imposed by IMPALA. In Opendss, the tap settings are by default ranging from 0.9 to 1.1 per unit (p.u.), divided into 32 distinct steps. This figure highlights the conversion of per unit values into corresponding discrete tap numbers, with the zero position being aligned with 0.9 p.u. Additionally, all taps are controlled to be at lower positions as to prevent the overvoltage violations in the studied case.
Fig. \ref{fig:system_components_eval1}.c vividly illustrates the contribution of each of the 30 PV systems in response to the solar irradiance profile of Santa Clara over a single day. This depiction also elucidates each PV system's contribution of injected power according to its location placed within the distribution network.
It is worth to mention that on a single machine, IMPALA encounters limitations with high core utilization, often leading to crashes or unresponsiveness. This constraint prevented us from running IMPALA on 24 cores like other algorithms, a scenario we aim to explore in future work on a higher-specification machine. Even with 12 cores, IMPALA is able to converge faster, a performance that is anticipated to be significantly enhanced when deploying on 24 cores. Furthermore, we conduct direct comparisons between our DRL approaches and three established model-based nad model-free methods: Model Predictive Control (MPC), Particle Swarm Optimization (PSO), and brute-force iterative sampling. These comparisons are aimed at evaluating each method's efficacy in achieving zero voltage violations and justifying the accuracy of the DRL method. Our findings reveal that while MPC lacks the assurance of completely eliminating voltage violations as illustrated in our previous work \cite{selim2023adaptive}, PSO requires a minimum of 100 particles and may reach zero violations but without guaranteed success for each iterative run. The brute-force method, though capable of identifying the optimal control action space, is significantly hampered by its computational intensity and the unpredictability of its time frame to achieve zero violations. This comparative analysis underscores the limitations of state-of-art methods in the real-time VVO problem.
\begin{table}[H]
\caption{Comparative Performance Analysis of DRL Architectures}
\centering
\begin{tabular}{lcccc}
\hline 
Architecture & CPUs & GPUs & \multicolumn{2}{c}{Performance} \\
\hline 
 & & & Reward & Time (H:M:S) \\
\hline 
Single-Machine & & & & \\
\hline 
SAC & 6 & 0 & 0 & 1:24:21\\
SAC & 12 & 0 & 0 & 0:42:15 \\
SAC & 24 & 0 & 0 & 0:24:19 \\
SAC & 24 & 1 & 0 & 0:20:31 \\
PPO & 6 & 0 & -0.262 & 0:12:35\\
PPO & 12 & 0 & -0.658 & 0:09:33 \\
PPO & 24 & 0 & -0.145 & 0:07:32 \\
PPO & 24 & 1 & -0.141 & 0:07:25 \\
IMPALA & 6 & 0 & 0 & 0:06:52 \\
IMPALA & 12 & 0 & \textbf{0} & \textbf{0:03:54} \\
\hline
\end{tabular}
\end{table}
\vspace{-18 pt}

\section{Conclusion}
In this investigation, we conducted a thorough examination of the effects of widespread DER integration, utilizing the IEEE 123-bus system with solar irradiance and load profiles based on California conditions. Leveraging the capabilities of the RLlib-IMPALA framework using the distributed computing and hyperparameter tuning, we achieved not only a rapid training process but also superior convergence rates in contrast to established methodologies. Moreover, this proposed framework is anticipated to significantly expedite the training process for many of DRL problems in the power system domain, which traditionally demand extensive durations to achieve convergence. However, the study encountered challenges, particularly when the IMPALA algorithm was subjected to rigorous core demands on an individual machine, necessitating meticulous hyperparameter tuning. In future work, we aim to evaluate the algorithm's performance on machines with an increased number of computational cores. Furthermore, a pivotal aspect of our research will be the exploration of VVO applications in large systems having more than 5000-nodes.

\bibliographystyle{ieeetr}
\bibliography{Mendeley.bib}

\begin{thebibliography}{10}

\bibitem{cao2020reinforcement}
D.~Cao, W.~Hu, J.~Zhao, G.~Zhang, B.~Zhang, Z.~Liu, Z.~Chen, and F.~Blaabjerg, ``{Reinforcement learning and its applications in modern power and energy systems: A review},'' {\em Journal of modern power systems and clean energy}, vol.~8, no.~6, pp.~1029--1042, 2020.

\bibitem{ma2023simplified}
Q.~Ma and C.~Deng, ``{Simplified Deep Reinforcement Learning Based Volt-Var Control of Topologically Variable Power System},'' {\em Journal of Modern Power Systems and Clean Energy}, 2023.

\bibitem{fan2022powergym}
T.-H. Fan, X.~Y. Lee, and Y.~Wang, ``Powergym: A reinforcement learning environment for volt-var control in power distribution systems,'' in {\em Learning for Dynamics and Control Conference}, pp.~21--33, PMLR, 2022.

\bibitem{liu2022bi}
H.~Liu, W.~Wu, and Y.~Wang, ``{Bi-level off-policy reinforcement learning for two-timescale volt/var control in active distribution networks},'' {\em IEEE Trans. Power Systems}, vol.~38, no.~1, pp.~385--395, 2022.

\bibitem{gupta2023deep}
S.~Gupta, A.~Mehrizi-Sani, S.~Chatzivasileiadis, and V.~Kekatos, ``{Deep Learning for Scalable Optimal Design of Incremental Volt/VAR Control Rules},'' {\em IEEE Control Systems Letters}, 2023.

\bibitem{manna2022learning}
S.~Manna, T.~D. Loeffler, R.~Batra, S.~Banik, H.~Chan, B.~Varughese, K.~Sasikumar, M.~Sternberg, T.~Peterka, M.~J. Cherukara, {\em et~al.}, ``{Learning in continuous action space for developing high dimensional potential energy models},'' {\em Nature communications}, vol.~13, no.~1, p.~368, 2022.

\bibitem{ming2023cooperative}
F.~Ming, F.~Gao, K.~Liu, and C.~Zhao, ``{Cooperative modular reinforcement learning for large discrete action space problem},'' {\em Neural Networks}, vol.~161, pp.~281--296, 2023.

\bibitem{williams2022trajectory}
K.~R. Williams, R.~Schlossman, D.~Whitten, J.~Ingram, S.~Musuvathy, J.~Pagan, K.~A. Williams, S.~Green, A.~Patel, A.~Mazumdar, {\em et~al.}, ``{Trajectory planning with deep reinforcement learning in high-level action spaces},'' {\em IEEE Trans. Aerospace and Electronic Systems}, 2022.

\bibitem{kanervisto2020action}
A.~Kanervisto, C.~Scheller, and V.~Hautam{\"a}ki, ``{Action space shaping in deep reinforcement learning},'' in {\em 2020 IEEE conference on games (CoG)}, pp.~479--486, IEEE, 2020.

\bibitem{raylibraries}
{Ray Team}, ``{Ray Documentation | Libraries}.'' \url{https://www.ray.io/libraries}, 2023.
\newblock Accessed: 2023-10-01.

\bibitem{anyscale}
{Anyscale Team}, ``{Ray Distributed Computing - Anyscale}.'' \url{https://www.anyscale.com/ray-distributed-computing}, 2023.
\newblock Accessed: 2023-10-01.

\bibitem{espeholt2018impala}
L.~Espeholt, H.~Soyer, R.~Munos, K.~Simonyan, V.~Mnih, T.~Ward, Y.~Doron, V.~Firoiu, T.~Harley, I.~Dunning, {\em et~al.}, ``{Impala: Scalable distributed deep-rl with importance weighted actor-learner architectures},'' in {\em International conference on machine learning}, pp.~1407--1416, PMLR, 2018.

\bibitem{haarnoja2018soft}
T.~Haarnoja, A.~Zhou, K.~Hartikainen, G.~Tucker, S.~Ha, J.~Tan, V.~Kumar, H.~Zhu, A.~Gupta, P.~Abbeel, {\em et~al.}, ``Soft actor-critic algorithms and applications,'' {\em arXiv preprint arXiv:1812.05905}, 2018.

\bibitem{schulman2017proximal}
J.~Schulman, F.~Wolski, P.~Dhariwal, A.~Radford, and O.~Klimov, ``Proximal policy optimization algorithms,'' {\em arXiv preprint arXiv:1707.06347}, 2017.

\bibitem{montenegro2012real}
D.~Montenegro, M.~Hernandez, and G.~Ramos, ``{Real time OpenDSS framework for distribution systems simulation and analysis},'' in {\em 2012 Sixth IEEE/PES Transmission and Distribution: Latin America Conference and Exposition (T\&D-LA)}, pp.~1--5, IEEE, 2012.

\bibitem{selim2023adaptive}
A.~Selim, J.~Zhao, F.~Ding, F.~Miao, and S.-Y. Park, ``{Adaptive Deep Reinforcement Learning Algorithm for Distribution System Cyber Attack Defense With High Penetration of DERs},'' {\em IEEE Trans. Smart Grid}, 2023.

\end{thebibliography}

\end{document}